\documentclass[10pt,english,aps,showkeys,notitlepage,nofootinbib,longbibliography]{revtex4-1}
\pdfoutput=1
\usepackage{color}
\usepackage[dvipsnames]{xcolor}
\usepackage{amsmath}
\usepackage{mathtools}
\usepackage{amssymb,amsthm}
\usepackage{graphicx}
\usepackage{esint}
\usepackage{babel}
\usepackage[normalem]{ulem} 
\usepackage{float}	
\usepackage[ruled,vlined]{algorithm2e}
\usepackage{lipsum}
\begin{document}

\def\myd{\textrm{d}}
\def\mye{\textrm{e}}
\newcommand{\norm}[1]{\lVert#1\rVert}


\title{Dense Limit of the Dawid-Skene Model
  for Crowdsourcing and Regions of Sub-optimality of Message Passing Algorithms}

\author{Christian Schmidt$^{1}$}
\affiliation{
$^1$ Institut de Physique Th\'eorique, CEA Saclay and CNRS, 91191, Gif-sur-Yvette, France.
}

\author{Lenka Zdeborov\'a$^{1}$}
\affiliation{
$^1$ Institut de Physique Th\'eorique, CEA Saclay and CNRS, 91191, Gif-sur-Yvette, France.
}

\begin{abstract}
Crowdsourcing is a strategy to categorize data through the contribution of many individuals. 
A wide range of theoretical and algorithmic contributions are based on the 
model of Dawid and Skene \cite{DawidSkene79}.
Recently it was shown in \cite{ok16,ok2016optimal} that, in certain regimes, 
belief propagation is asymptotically optimal for data generated from the Dawid-Skene model. 
This paper is motivated by this recent progress.
We analyze the dense limit of the Dawid-Skene model. 
It is shown that it belongs to a larger class of low-rank
matrix estimation problems for which it is possible to express the asymptotic, 
Bayes-optimal, performance in a simple closed form. 
In the dense limit the mapping to a low-rank matrix estimation problem 
provides an approximate message passing algorithm that solves the problem algorithmically.
We identify the regions where the algorithm efficiently computes the Bayes-optimal estimates.
Our analysis refines the results of \cite{ok16,ok2016optimal} about optimality of message
passing algorithms by characterizing regions of parameters where these
algorithms do not match the Bayes-optimal performance. We further
study numerically the performance of approximate message passing, 
derived in the dense limit, on sparse instances 
and carry out experiments on a real world dataset.
\end{abstract} 

\maketitle

\section{Introduction \label{sec:introduction}}

The development of large-scale crowdsourcing platforms, such as Amazon's MTurk, has
popularized crowdsourcing as a simple approach to solve various
problems that remain difficult for computers but require 
little effort to human workers. The overall strategy is simple: the
requester poses a set of tasks that are allocated to several
individuals from a pool of workers (the crowd). The workers answer
according to their abilities and their will. Importantly, the set of
answers is typically not unambiguous and post-processing has to be
performed in order to infer the true information (typically labels) from the noisy
observations (answers). 
With the crowds answers at hand the objective becomes to infer the true labels 
with as few mistakes as possible.
The outcome of such a strategy strongly depends on the competences of the individuals; 
which makes it necessary to infer not only the true labels, but also the competences of the individuals.

A large fraction of the theoretical work on crowdsourcing
focuses on the so-called Dawid-Skene (DS) model, after the authors of the
seminal paper \cite{DawidSkene79}. 
In the DS model we consider $N$ workers, each of them of a certain
reliability that denotes the probability that a worker gives the correct answer, 
represented by $0 \le p^0_{i} \le 1$ for worker
$i=1,\dots,N$. Further there are $M$ tasks, each having a true label that we
denote by $v^0_j \in \{\pm 1\}$ for task $j=1,\dots,M$. The worker $i$ is
assigned a subset of tasks $j \in \partial i$ 
to which it assigns an answer $Y_{ij} \in \{\pm1\}$. 
We denote $Y_{ij}=0$ if $j\notin \partial i$, that is  
for tasks $j$ that were not assigned to worker $i$. In the DS model
labels provided by worker $i$ for task $j$ are modeled as 
\begin{equation}
       P(Y_{ij} ) =   p^0_{i}  \delta( Y_{ij} - v^0_j  ) +
       (1-p^0_{i})  \delta( Y_{ij} + v^0_j  )\, .
\end{equation}
Moreover is it assumed that the $p^0_i$s are drawn
independently from some probability distribution $P_{p^0}$. 

The task allocation design (which tasks gets assigned to which worker) is in general part of the
crowdsourcing problem and various strategies have been described and
studied in the literature. It has been argued that
designing the graph of assignments at random has
practical and optimality advantages, among others it enables a sharp
theoretical analysis of the problem, see e.g.~\cite{KargerOhShah11}. On bipartite random regular graphs,
where every worker is assigned $r$ tasks and every task is assigned to
$l$ workers, the DS model has been studied in detail by \cite{KargerOhShah11,ok16}. 

While in general reconstructing the true labels and workers reliabilities from
the observed answers $Y_{ij}$ is an NP-hard problem, the authors of \cite{ok16,ok2016optimal}
obtained a remarkable theorem stating that in certain
regions of parameters belief propagation reconstructs
the true labels optimally in the limit of large system sizes. The
belief propagation algorithm for crowdsourcing was first suggested by
\cite{Liu12}. Many other algorithms for crowdsourcing exist in the
literature, but as far as we know none of them reaches optimal
performance for large random instances of the DS model for a
regime where the probability of error per task stays bounded away
from zero. 

The goal of the present paper is to carry out an asymptotic analysis 
of the DS model in the dense regime where each worker is 
assigned a constant fraction of the $M$ tasks.
Otherwise we are in the same setting as \cite{ok16,ok2016optimal}, i.e. with
random worker reliabilities and on random graphs.
From our analysis it is possible to characterize more
tightly the region of parameters for which belief propagation is
optimal and for which it is not. We find cases where a first
order phase transition appears in the error of reconstruction of the
true labels. Such a first order phase transition is 
associated with a region of parameters in which belief propagation does
not match the asymptotically optimal performance. Our work can thus
be seen as a follow-up on \cite{ok16,ok2016optimal} providing a
refined analysis of the regions of parameters for which 
belief propagation is or is not asymptotically optimal.

In section \ref{sec:model} we first define a dense version of the
DS model. In the dense DS model workers reliabilities are close to $1/2$ as otherwise inference becomes trivially easy. 
The dense DS model belongs to a class of low-rank matrix factorization problems, as studied
recently by statistical physics techniques in 
\cite{LesieurKrzakalaZdeborova15,LesieurKrzakalaZdeborova17}. 
The authors derived the approximate message passing (AMP) algorithm 
that efficiently computes the Bayes-optimal estimator and 
analyzed the Bayes-optimal performance in a closed form.
One of the merits of AMP is that its
asymptotic performance can be described via the so-called state
evolution, as proven in \cite{rangan2012iterative,javanmard2013state}. 
The performance of the Bayes-optimal estimator was also later put on fully
rigorous bases in the work of \cite{MiolaneUV17} under assumptions
that include the dense DS model as considered in
section \ref{sec:model}. We apply the results derived in those papers
and identify the region of parameters for which the associated
approximate message passing algorithm is suboptimal. In
section \ref{sec:sparse_limit} we then investigate numerically how the results -- valid
in the dense regime -- transfer into the sparse regime as originally considered in \cite{ok16,ok2016optimal}. 
We finally carry out some experiments on a real world dataset 
to show that AMP reaches similar performance to competing state-of-the-art algorithms.

\section{Dense Limit of the Dawid-Skene Model \label{sec:model}}

\subsection{Definition of the dense limit \label{subsec:def}}

In this section the dense Dawid-Skene (dDS) model for
crowdsourcing is introduced where each of the $N$ workers 
is assigned a constant fraction of the $M$ questions.  
It is shown that it can be modeled as a low-rank 
matrix factorization problem as studied in 
\cite{DeshpandeM14,NIPS2013_5074,LesieurKrzakalaZdeborova15,LesieurKrzakalaZdeborova17,MiolaneUV17}.

Let the probability that worker $i$ provides a correct answer, $p^0_i$, be close to $1/2$ and introduce the
parameters $\nu$ and $\theta^0_{i}$ such that $p^0_i = (1 + \sqrt{\nu/ N} \,
\theta^0_i )/2 $. The parameter $\nu$ is an overall scale parameter,
while $\theta^0_{i}$ is the rescaled reliability of worker $i$ taken
from some probability distribution $P_{\theta^0}$.  
In the dense limit the $1/\sqrt{N}$ scaling causes all the
reliabilities to be close to $1/2$ which is the interesting regime 
because a total of $\Theta(N)$ answers is received for each question.\footnote{
We make use of the standard big-theta and big-O notation. We refer to a
function as $\Theta(N)$ if its dominant asymptotic growth rate is proportional to $N$.
While $O(N)$ refers to an asymptotic growth rate that is bounded by some constant times $N$.
}
If we had another scaling, the problem would become either trivially
hard or easy in the thermodynamic limit, $N\to \infty$.

The true label of task $j$ is $v^0_j$, with $v^0_j \in \{\pm
1\}$ distributed as $P_{v^0}$. 
We denote with $Y_{ij}$ the label assigned to question $j$ by
worker $i$ and assume $Y_{ij} \in \{0,\pm1\}$. If
$Y_{ij}=0$ question $j$ was left out by worker $i$. Set $\alpha \coloneqq \frac{M}{N}$ and
consider a system in which each worker is posed $(1-\rho)M$
questions in the limit where $M,N \to \infty$, while $\alpha=\Theta(1)$
and $\rho=\Theta(1)$. In
this limit the likelihood in the DS model becomes 
	\begin{align}
	\begin{split}
	P\left( Y_{ij} = \pm 1 \mid \theta_i ,\  v_j \right) &= (1-\rho) \cdot \frac{1}{2} \cdot \left( 1 \pm \sqrt{\frac{\nu}{N}} \, \theta_i v_j\right)
	\\
	P\left( Y_{ij} = 0 \mid \theta_i ,\  v_j \right) &= \rho
	\end{split}
	\, ,
	\label{eq:channel}
	\end{align}
	
where we assumed that the fraction of un-answered questions, $\rho$,
is independent of $(i,j)$. 
The rest of the present section is set in the limit where $N\to
\infty$ and all the other parameters $\theta_i, \nu, \alpha, \rho =
\Theta(1)$.
Later, in section \ref{sec:sparse_limit}, we will discuss how to
extrapolate the results into the sparse regime where each worker is
only assigned to $O(1)$ tasks.

Worker with $\theta_i = 0$ give answers that are completely uninformative and will be
called \emph{spammers}. On the contrary, if $\theta_i \gg 1$ 
the answers are ``strongly'' aligned with the truth and 
we refer to such workers as \emph{hammers}. 
Adversaries are characterized by $\theta_i< 0$. They are also considered hammers if
$\theta_i \ll -1$ because their answers are aligned against the truth, 
as opposed to the random alignment of the spammers.

\subsection{Equivalence to low-rank matrix estimation \label{subsec:low_rank}}

The dDS model is a special case of bipartite low-rank (rank one in the present case) matrix factorization
as formulated in a much more general setting in \cite{LesieurKrzakalaZdeborova17}. 
In the rest of this section we follow closely that paper and
review the results that will be applied to the present model.

In the theoretical part of this work (i.e.~in all but section
\ref{subsec:real_world}) we assume that the distributions from which
the ground truth reliabilities $\theta_i^0$ and labels $v^0_j$ are drawn,
$P_{\theta^0}$ and $P_{v^0}$ respectively, are known. Under these assumptions 
we aim to (a) compute efficiently the Bayes-optimal estimators 
of $\theta_i^0$ and $v^0_j$, given the answers $Y_{ij}$ and (b) to 
evaluate the asymptotic inference performance. 

Denoting by $\boldsymbol{\theta} \in {\mathbb R}^N$ the vector of rescaled reliabilities for all $N$
workers, and $\mathbf{v}\in {\mathbb R}^M$ the vector of labels, we set
	\begin{equation}
	\mathbf{w} \coloneqq \frac{\boldsymbol{\theta}\,\mathbf{v}^T}{\sqrt{N}}
	\label{eq:low_rank_matrix}
	\end{equation}
and re-express (\ref{eq:channel}) as 
	\begin{equation}
	P\left( Y_{ij} \mid w_{ij} \right) = \exp\left( g(Y_{ij},w_{ij}) \right)
	\hspace{0.5cm} , \hspace{0.5cm} 
	g(Y_{ij},w_{ij}) = 
	\begin{cases} 
	\log\left( \frac{(1-\rho)}{2} \right) + \log \left(1\pm \sqrt{\nu}w_{ij}\right) & \textrm{if }Y_{ij}=\pm 1 
	\\ 
	\log\left(\rho\right) & \textrm{if } Y_{ij}=0
	\end{cases}
	\, .
	\label{eq:g}
	\end{equation}

From Bayes' theorem we obtain the corresponding posterior probability distribution
	\begin{equation}
	P( \boldsymbol{\theta}, \mathbf{v} \mid \mathbf{Y} ) = \frac{1}{Z(\mathbf{Y})} 
	\prod_{1\leq i \leq N} P_{\theta}(\theta_i)
	\prod_{1\leq j \leq M} P_{v}(v_j)
	\prod_{1\leq i \leq N, 1\leq j \leq M} \textrm{e}^{g(Y_{ij},w_{ij})}
	\, .
	\label{eq:posterior}
	\end{equation}
The Bayes-optimal estimates, $\hat{\boldsymbol\theta}$, that minimize the mean-squared-error (MSE) on $\boldsymbol{\theta}$
	\begin{equation}
	\text{MSE}_\theta = \frac{1}{N} \sum_i \left( \hat\theta_i - \theta^0_i \right)^2
	\label{eq:def_mse}
	\end{equation}
and the bitwise error-rate (ER) on $\mathbf{v}$
	\begin{equation}
	\text{ER}_v 
	= 
	\frac{1}{M} \sum_{j} \mathbb{I}\left[ \hat{v}_j \neq v^0_j \right] 
	= 
	\frac{1}{M} \sum_{j} \left( \frac{\hat{v}_j - v^0_j}{2} \right)^2 
	=
	\frac{1}{2} \frac{1}{M} \sum_{j} \left( 1- \hat{v}_j v^0_j \right)
	\label{eq:def_er}
	\end{equation}
read
\begin{equation}
	\hat{\theta}^{\text{MMSE}}_{i}(\mathbf{Y}) = \int \myd \theta_{i} \theta_{i} P \left(  \theta_{i} \mid \mathbf{Y} \right)
	\hspace{0.5cm}
	\text{and}
	\hspace{0.5cm}
	\hat{v}^{\text{MER}}_{j}(\mathbf{Y}) = \text{sign} \int \myd v_{j} v_{j} P \left(  v_{j} \mid \mathbf{Y} \right)
	\, , \label{Bayes_optimal}
\end{equation}
where $P \left(x_{k} \mid \mathbf{Y} \right)$, with $x_k\in
\{\{\theta\}_{i=1,\dots, N},\{v_j\}_{j=1,\dots,M}\}$, is the posterior marginal of $(\ref{eq:posterior})$ after integrating out all variables except $x_k$.
Hence inferring the reliabilities and labels in the crowdsourcing problem 
reduces to evaluating the marginal expectations of the 
posterior probability distribution. In general this is a difficult task. 
The contribution of the present work is to realize that the 
dDS model falls into a class of low-rank matrix estimation problems 
for which the posterior probability distribution can be evaluated, as shown in \cite{LesieurKrzakalaZdeborova17}. 
Using these results, the phase diagram can be evaluated in great detail.

\subsection{Approximate message passing \label{sec:AMP}}

The approximate message passing (AMP) algorithm for low-rank matrix estimation 
is a simplification of belief propagation in the limit of
dense graphical models. In this limit both, belief propagation and AMP 
have the same asymptotic performance. However, AMP is much simpler to
implement and has a favorable scaling w.r.t~the problem size. It is closely related to the
Thouless-Anderson-Palmer equations \cite{ThoulessAnderson77} from the theory of spin
glasses, with correct time indices \cite{bolthausen2014iterative,zdeborova2015statistical}. 
AMP for low-rank matrix factorization was first derived for special cases in
\cite{rangan2012iterative,NIPS2013_5074} and in its general form in
\cite{LesieurKrzakalaZdeborova15,LesieurKrzakalaZdeborova17}. 

AMP can be derived starting from belief propagation for the graphical model where both
the reliabilities and labels are variable nodes, and there are
pair-wise factor nodes corresponding to the answers $Y_{ij}$. 
The following two simplifications of BP are then made. First, the BP
messages are replaced by their means and variances
which eradicates the necessity of tracking a whole function for each
message. Secondly, each (mean and variance) \emph{message} is replaces by its
\emph{marginal} version, reducing the complexity from $O(N^2)$ messages to
$O(N)$ marginals. For details we refer the reader to
\cite{LesieurKrzakalaZdeborova17}.

To state the AMP algorithm for the dense DS model it is
necessary to specify the denoising functions
$f_{\theta}(A_{\theta},B_{\theta})$ and $f_v(A_v,B_v)$ that depend on
the priors $P_\theta$ and $P_v$ respectively. 
$A$ and $B$ are estimates for the parameters of a Gaussian distribution
that are computed self-consistently. 
The estimate $\hat x_k$ 
-- with $x_k \in \{ \{ \theta_i\}_{i=1,\dots,N}, \{v_j\}_{j=1,\dots,M} \}$ -- 
are then computed as the mean of the prior weighted with this effective Gaussian. 
The estimates for their variance are obtained from the derivative w.r.t.~$B$.
	\begin{eqnarray}
	\hat{x} \coloneqq f_x(A_x,B_x) = \frac{1}{Z_x(A_x,B_x)} \int \myd x \, x \, P_x(x) \, \mye^{- \frac{1}{2} A_x x^2 + B_x x}
	\ ,
	\hspace{1cm}
	\sigma_x  = \partial_{B_x} \, f_x(A_x,B_x)
	\, ,
	\label{eq:denoising_function}
	\end{eqnarray}
where $x$ can stay either for $\theta$ or $v$. 

To state AMP we need to define the Fisher score matrix  
	\begin{eqnarray}
	S_{ij} & \coloneqq & \left. \frac{\partial g( Y_{ij}, w_{ij} )}{\partial w_{ij}} \right\vert_{w_{ij}=0} = Y_{ij}\cdot \sqrt{\nu}
	\, ,
	\label{eq:R_S}
	\end{eqnarray}
where $g( Y_{ij}, w_{ij} )$ is defined in eq.~(\ref{eq:g}). 
Further we define the Fisher information (inverse effective noise) of
the noisy observation channel 
	\begin{equation}
	\Delta^{-1} = \mathbb{E}_{P(Y_{ij} \mid w_{ij}=0)} \left[ \left( \left. \frac{\partial g( Y_{ij}, w_{ij} )}{\partial w_{ij}} \right\vert_{w_{ij}=0} \right)^2 \right] = (1-\rho)\nu
	\, .
	\label{eq:effective_noise}
	\end{equation}
	
Given these definitions AMP is an iterative scheme that we outline in
Algorithm~\ref{alg:AMP}. The numerical implementation might profit
from an adequate damping scheme in order to enhance convergence even
on small instances or when the model assumptions are not satisfied.

\begin{algorithm}[H] 
	\SetAlgoNoLine
		\KwData{
		$\mathbf{S},\ \Delta, \  \delta $
		\tcp*{$\mathbf{S}$ and $\Delta$ according to (\ref{eq:R_S}) and (\ref{eq:effective_noise}) respectively.}
		}

		\KwResult{MMSE estimates $\hat{\mathbf{v}}$ and $\hat{\boldsymbol\theta}$ }

		Initialize: 
		$\hat{\mathbf{v}} \gets \hat{\mathbf{v}}^{\textrm{init}} \sim P_v(\mathbf{v})$,  
		$\hat{\boldsymbol\theta} \gets \hat{\boldsymbol\theta}^{\textrm{init}} \sim P_\theta(\boldsymbol{\theta})$ ;
		$\sigma_v \gets 1$,	$\sigma_\theta \gets 1 $;
		$\hat{\mathbf{v}}^{\textrm{old}} \gets \mathbf{0}$,  $\hat{\boldsymbol\theta}^{\textrm{old}} \gets \mathbf{0}$ 
		\;

		\While{ $\norm{\hat{\boldsymbol\theta} - \hat{\boldsymbol\theta}^{\textrm{old}}}_2^2 + \norm{\hat{\mathbf{v}} - \hat{\mathbf{v}}^{\textrm{old}}}_2^2 > \delta$}
			{
	
			$\mathbf{B}_\theta \gets \frac{1}{\sqrt{N}} \, \mathbf{S} \hat{\mathbf{v}} - \frac{1}{\Delta} \, \hat{\boldsymbol{\theta}}^{\textrm{old}} \sigma_v$ \;
		
			$\mathbf{A}_\theta \gets \frac{1}{N \Delta} \, \hat{\mathbf{v}}^T \hat{\mathbf{v}}$ \;
		
			$\mathbf{B}_v \gets \frac{1}{\sqrt{N}} \, \mathbf{S}^T \hat{\boldsymbol{\theta}} - \frac{\alpha}{\Delta} \, \hat{\mathbf{v}}^{\textrm{old}} \sigma_\theta$ \;
			
			$\mathbf{A}_v \gets \frac{1}{N \Delta} \, \hat{\boldsymbol{\theta}}^T \hat{\boldsymbol{\theta}}$ \;

			$\hat{\boldsymbol{\theta}}^{\textrm{old}} \gets \hat{\boldsymbol{\theta}}$,   $\hat{\mathbf{v}}^{\textrm{old}} \gets \hat{\mathbf{v}}$ \;
			
			$\hat{\boldsymbol{\theta}} \gets f_{\theta}(\mathbf{A}_{\theta},\mathbf{B}_{\theta})$, $\sigma_\theta \gets \frac{1}{N} \sum_{1\leq i\leq N}\partial_{B_{\theta_i}} \,f_{\theta}(A_{\theta_i},B_{\theta_i})$ \;
			
			$\hat{\mathbf{v}} \gets f_{v}(\mathbf{A}_{v},\mathbf{B}_{v})$, $\sigma_v \gets \frac{1}{M} \sum_{1 \leq j \leq M} \partial_{B_{v_j}} \,f_{v}(A_{v_j},B_{v_j})$ \;
			
			}

		\caption{Approximate message passing for crowd sourcing.}
		\label{alg:AMP}
	\end{algorithm}

\subsection{State Evolution  \label{sec:state_evolution}}

The AMP algorithm depends on the realization of the disorder
$\mathbf{Y}$ and consequently so do the AMP estimates
$\hat{\boldsymbol{\theta}}, \ \hat{\mathbf{v}}$ for the 
reliabilities and task labels. Quite remarkably, in the large size
limit $N\to \infty$, the performance of the algorithm can be tracked
with high probability by the so-called state evolution (SE) equations. This
has been proven rigorously in \cite{rangan2012iterative,javanmard2013state}.

In the Bayes-optimal setting, where the true distributions $P_{\theta^0}$ and $P_{v^0}$ are known and equal to $P_{\theta}$ and $P_{v}$ respectively, the overlap of the AMP estimates with the true solution can be quantified in terms of the two order parameters
	\begin{align}
	\begin{split}
	M^t_\theta 	&= \frac{1}{N} \sum_{1\leq i \leq N}
        \hat{\theta}^t_i \, \theta^0_{i}\, , \\
	M^t_v 		&= \frac{1}{M} \sum_{1\leq j \leq M} \hat{v}^t_j \, v^0_{j}
	\, .
	\end{split}
	\label{eq:order_parameters}
	\end{align}
Where $x^0$ indicates the true value of $x$, and $t$ the iteration step of the AMP equations (Alg.~\ref{alg:AMP}).

The SE equations imply that these order parameters evolve with high probability as 
	\begin{align}
	\begin{split}
	M_v^{t+1} &= \mathbb{E}_{v^0,W}  \left[  f_v \left(
            \frac{M_\theta^t}{\Delta} \, , \,
            \frac{M_\theta^t}{\Delta} \, v^0 +
            \sqrt{\frac{M_\theta^t}{\Delta}} \, W  \right) \, v^0
        \right]\, ,
	\\
	M_\theta^{t} &= \mathbb{E}_{\theta^0,W}  \left[ f_\theta \left( \frac{\alpha \, M_v^t}{\Delta} \, , \, \frac{\alpha \, M_v^t}{\Delta} \, \theta^0 + \sqrt{\frac{\alpha \, M_v^t}{\Delta}} \, W \right) \, \theta^0 \right]
	\, .
	\end{split}
	\label{eq:state_evolution}
	\end{align}
where $W$ is an effective Gaussian random variable of zero mean and
unit variance, $v^0 \sim P_v$, $\theta^0 \sim P_\theta$, the
functions $f_v$ and $f_\theta$ are defined in (\ref{eq:denoising_function}), $\alpha = M/N$
and $\Delta$ is the effective noise (\ref{eq:effective_noise}). 

Let us call $M^{\rm SE}_\theta$ and $M^{\rm SE}_v$ the fixed points of
the SE equations (\ref{eq:state_evolution}). 
These fixed points are then
associated to the MSE (\ref{eq:def_mse}) and ER (\ref{eq:def_er}) as
reached by the AMP algorithm through 
	\begin{eqnarray}
	\text{MSE}^{\rm AMP}_{\theta} &=& {\mathbb E}_{\theta} (\theta^2 ) -
                                M^{\rm SE}_\theta \, ,
	\label{eq:MSE}
	\\
	\text{ER}^{\rm AMP}_{v} &=& (1 - R^{\rm SE}_v)/2
	\label{eq:ER}
	\, ,
	\end{eqnarray}
where we introduced	the order parameter $R^t_v = 1/M \sum_{i} \text{sign}(\hat{v}^t_i) \, v^0_{i}$
	\begin{equation}
	R^{\rm SE}_v = \mathbb{E}_{v^0,W}  \left\{  \text{sign} \left[ f_v
            \left( \frac{M^{\rm SE}_\theta}{\Delta} \, , \,
              \frac{M^{\rm SE}_\theta}{\Delta} \, v^0 +
              \sqrt{\frac{M^{\rm SE}_\theta}{\Delta}} \, W  \right) \right] \,
          v^0 \right\} 
          \, .
          \label{eq:RRR}
	\end{equation}

\subsection{Bayes-optimal error and sub-optimality of message passing
  algorithms \label{sec:Bayes-optimal}}

As conjectured in \cite{LesieurKrzakalaZdeborova17} and proven
rigorously in \cite{MiolaneUV17} the performance of the  
Bayes-optimal estimator (\ref{Bayes_optimal}) can be evaluated in the
large size limit $N\to \infty$ with $\alpha=\Theta(1)$ from the global minimizer
of the so-called replica symmetric Bethe free energy, which reads 
	\begin{eqnarray}
	\phi_{\textrm{Bethe}}(M_\theta,M_v) &=& 
	\alpha \frac{M_\theta M_v}{2\Delta} - 
	\alpha \mathbb{E}_{v^0,W} \left[ \log Z_v \left(
          \frac{M_\theta}{\Delta} \, , \, \frac{M_\theta}{\Delta}
          \, v^0 + \sqrt{\frac{M_\theta}{\Delta}} \, W \right)
          \right] -  \nonumber  \\
	&& - \, \mathbb{E}_{\theta^0,W} \left[ \log Z_\theta \left( \frac{\alpha \, M_v}{\Delta} \, , \, \frac{\alpha \, M_v}{\Delta} \, \theta^0 + \sqrt{\frac{\alpha \, M_v}{\Delta}} \, W \right) \right]
	\, .
	\label{eq:free_energy}
	\end{eqnarray}
where the functions $Z_\theta$ and $Z_v$ are defined in
(\ref{eq:denoising_function}) and the rest of the variables are
defined in the same way as in the SE. Assume $M^*_\theta$ and $M^*_v$ are the
global minimizers of the above Bethe free energy. Then the
minimum-mean-squared-error (MMSE) and the minimum-error-rate (MER) are
expressed as 
	\begin{eqnarray}
	\text{MMSE}_{\theta} &=& {\mathbb E}_{\theta}  (\theta^2) -
                                M^*_\theta \, ,
	\label{eq:MMSE}
	\\
	\text{MER}_{v} &=& \frac{1}{2} ( 1 - R^*_v)
	\label{eq:MER}
	\, ,
	\end{eqnarray}
where $R^*_v$ is obtained from $M^*_\theta$ via (\ref{eq:RRR}).

It is straightforward to observe that the SE equations are
in fact stationarity conditions of the Bethe free energy. Hence the fixed
points of the state evolution are critical points of the Bethe free
energy. Whether or not the SE reaches the global minimizer
$M^*_\theta,\ M^*_v$ depends on the shape of the Bethe free
energy and the initialization of the SE
equations at $t=0$. Canonically the SE is initialized in
such a way that the initial estimators are simply taken from the
prior distributions.  

We can now explain the key point of the present paper. Previous
work \cite{ok16,ok2016optimal} proved asymptotic optimality of belief
propagation under certain assumptions on the parameters of the model. 
The present analysis of the dDS model is able to determine sharply in what
regions of parameters AMP matches the Bayes-optimal estimator and when
it does not, thus refining the previous picture in the limit, 
where AMP and BP are asymptotically equivalent. 

Previously we reduced the high-dimensional model 
into the investigation of the two-variable free energy function
(\ref{eq:free_energy}). 
In particular, the phases in which AMP does not match the 
Bayes-optimal estimator can be characterized in terms of 
the critical points of the free energy and whether or not the
state evolution (\ref{eq:state_evolution})
converges to the global minimum of the free energy (\ref{eq:free_energy}). 
The way we check this in practice is that we
initialize the state evolution in two different ways: 
\begin{itemize}
     \item{Uninformative initialization, where $M_v^{t=0}=( \mathbb
         E_{v} (v))^2$ and $M_\theta^{t=0}=(\mathbb E_{\theta}
         (\theta))^2$. This corresponds to the uninformative
         initialization of the algorithm where the initial values of
         the estimators are simply taken equal to the mean of the
         prior distributions $P_\theta$ and $P_v$. The error achieved by
         the AMP algorithm is then given by iteration of
         (\ref{eq:state_evolution}) from this uninformative initialization. }
     \item{Informative initialization, where $M_v^{t=0}=\mathbb
         E_{v} (v^2)$ and $M_\theta^{t=0}=\mathbb E_{\theta}
         (\theta^2)$ so that the initial mean-squared-errors are
         zero. This is not possible within the algorithm
         without the knowledge of the ground truth and it is purely used
         for the purpose of the analysis. If the iteration of the SE 
         equations (\ref{eq:state_evolution}) from this informative
       initialization leads to a different fixed point than from the
       uninformative initialization, then the free energies of the two
       fixed points need to be compared and the larger one
       surely does not correspond to the Bayes-optimal performance. }
\end{itemize}
This procedure is sufficient provided there are no other fixed points. If there are, the free
energy of all of them needs to be compared.

\paragraph*{Zero-mean priors and uninformative fixed point}
If both prior distributions $P_\theta$ and $P_v$ have zero
mean, the uninformative initialization $M_\theta=M_v=0$ is a
fixed point of the SE and equations
(\ref{eq:state_evolution}) can be expanded around this fixed point. 
In first order we obtain 
	\begin{eqnarray}
	M_\theta^{t} = \frac{\alpha}{\Delta} \left(\mathbb{E}_\theta \left[ \theta^2 \right]\right)^2 M_v^t
	\\
	M_v^{t+1} = \frac{1}{\Delta} \left(\mathbb{E}_v \left[ v^2 \right]\right)^2 M_\theta^t
	\, ,
	\end{eqnarray}
implying that the uninformative fixed point is numerically stable for $\Delta^2 >
\alpha \left(\mathbb{E}_v \left[ v^2 \right]\right)^2
\left(\mathbb{E}_\theta \left[ \theta^2 \right]\right)^2 $ and
unstable otherwise. Therefore we define the critical effective noise,
$\Delta_{\text{c}}$, as
	\begin{equation}
	\Delta_{\text{c}} = \sqrt{\alpha} \cdot \mathbb{E}_\theta \left[ \theta^2 \right] \mathbb{E}_v \left[ v^2 \right]
	\, .
	\label{eq:linear_stability}
	\end{equation}
For $\Delta < \Delta_{\text{c}}$ the uninformative initialization becomes
numerically unstable. 
The threshold $\Delta_{\text{c}}$
correspond to the 2nd order phase transition in the behaviour of the
AMP algorithm, meaning that the overlap reached by the algorithm is
non-analytic and continuous at $\Delta_{\text{c}}$.

In the case where both the priors, $P_\theta$ and $P_v$, have zero mean,
we can divide the region of parameters into the following three phases:
	\begin{itemize}
	\item \textbf{Easy phase:} The free energy (\ref{eq:free_energy})
          has a unique minimum and this minimum is associated
          with a positive overlap with the ground-truth configuration. Consequently
          iterating the state evolution (\ref{eq:state_evolution})
          yields an informative fixed point from both, the
          informative, as well as the (perturbed) uninformative initializations. 
          AMP is Bayes-optimal.
	\item \textbf{Hard phase:} In this phase at least two minima of the free
          energy (\ref{eq:free_energy}) coexist; at least one local minimum of
          small overlap and a global minimum of larger overlap. 
         The outcome of iterating the state evolution equations now
         depends on the initialization: while the informative
         initialization yields a fixed point with large overlap, the
         uninformative initialization leads to a fixed point of low overlap. 
         This is precisely the region of parameters where the
         approximate message passing algorithms do not reach the
         information-theoretically optimal performance and 
         \emph{AMP is not Bayes-optimal}.
	\item \textbf{Impossible phase:} When the global minimum
          of (\ref{eq:free_energy}) is associated to the trivial,
          non-informative, fixed point corresponding to zero overlap,
          we talk about a phase of impossible inference. 
          Otherwise this region is indeed similar to the easy phase in the 
          sense that AMP is Bayes-optimal.
	\end{itemize}
If at least one of the priors has non-zero mean, then the distinction 
of an impossible phase is not meaningful and one would only have an 
easy and a hard phase, the later is defined by asymptotic
sub-optimality of the AMP algorithm. 

Let us further define the following three thresholds that are associated with
the existence of a hard phase. The hard phase is always linked
to the presence of a first order phase transition, i.e. a
discontinuity in the asymptotic value of the overlap reached by the
Bayes-optimal estimator. The {\it algorithmic threshold} $\Delta_{\rm
alg}$ is the largest value of effective noise, $\Delta$, below which the AMP
algorithm asymptotically matches the Bayes-optimal performance.  The
{\it spinodal threshold}, $\Delta_{\rm sp}$, is the smallest values of
effective noise above which the informative initialization converges to a
different fixed point than the (perturbed) uninformative
initialization. The \emph{information theoretic transition}, $
\Delta_{\rm alg} < \Delta_{\rm IT} < \Delta_{\rm sp}$, is where 
the value of the Bethe free energy of the fixed
point reached from the uninformative initialization crosses with the
free energy of the fixed point reached from the informative
initialization. The discontinuity in overlap happens at $\Delta_{\rm IT}$. 
Remark that while in some models, such as the
stochastic block model \cite{LesieurKrzakalaZdeborova15}, we find
$\Delta_{\text{c}} = \Delta_{\rm alg}$ in general and in the present model $\Delta_{\text{c}} \neq \Delta_{\rm alg}$.

\section{Phase Diagrams for the dense David-Skene model\label{sec:phase_diagram}}

A key property of the results we described so far is that the
asymptotic behaviour of the AMP algorithm and of the Bayes-optimal
estimatior depend only on the priors $P_v$, $P_\theta$ and the
effective noise $\Delta = 1 / [(1-\rho) \nu]$. 
In what follows concrete priors will be considered.

It is assumed that the ground truth task labels are generated from 
	\begin{equation}
	P_v(v_j) = (1-\beta)\delta\left( v_j-1 \right) + \beta \delta \left( v_j+1 \right)
	\, .
	\label{eq:prior_v}
	\end{equation}	
With the parameter $\beta\in [0,1]$ accounting for a bias in the
dataset.

We start our discussion with worker reliabilities $\theta_i$ that were
drawn from a skewed Rademacher-Bernoulli (RB) prior
	\begin{equation}
	P_\theta (\theta) = \left(1-\mu\right) \delta(\theta) + \mu \left[ (1-\lambda)\delta(\theta -1) + \lambda \delta (\theta +1) \right]
	\, .
	\label{eq:rb_prior}
	\end{equation} 
Besides its simplicity the phase diagram for this case comprises the
essential features.
Tuning $\mu$ from zero to one interpolates between an uninformative
crowd of mere spammers and an informative crowd. 
The fraction of adversaries is controlled by $\lambda$. 
In physics terms the workers with $\theta=-1$ are spins that are 
coupled to the questions by an anti-ferromagnetic interaction, 
whereas the workers with $\theta=1$ are ferromagnetically coupled. 
Consequently also the adversaries enhance our ability to recover the 
correct labels, if they can be identified, as they align anti-parallel to the truth.

The RB prior is the dense version of what is sometimes
referred to as the ``spammer-hammer'' model in the literature
\cite{KargerOhShah11}: Workers are either spammers that provide random
answers or hammers that align very strongly with (or opposed to) the
truth. Here the situation is slightly different as we assume a very
weak alignment of $\Theta(1/\sqrt{N})$, cf. (\ref{eq:channel}). In the dDS
sending $\nu\to\infty$ and thus $\Delta\to 0$ approximates the hammers. 
The limit $\nu\to N$ will be considered in section \ref{sec:sparse_limit}.

\subsection{The case of symmetric priors}

If $\lambda=1/2$ and $\beta=1/2$ both the priors $P_v$ and $P_\theta$
have zero mean and the SE equations in (\ref{eq:state_evolution})
have a trivial fixed point at $M^*_v = M^*_\theta = 0$. Expansion around
this fixed point yields
	\begin{equation}
	M_v^{t+1} = \alpha \frac{\mu^2}{\Delta^2} \cdot M_v^t - \alpha^2 \frac{\mu^2}{\Delta^2} \left[\frac{\mu}{\Delta}+\frac{\mu^2}{\Delta^2}\right] \cdot \left(M_v^t\right)^2 + O((M_v^t)^3)
	\, .
	\label{eq:linear_RB}
	\end{equation}
The linear term gives the stability criterion of the trivial fixed point that we had already derived in (\ref{eq:linear_stability})
	\begin{equation}
         \Delta_{\textrm{c}} = \sqrt{\alpha} \cdot \mu 
	\, .
	\label{eq:stability_criterion_RB}
	\end{equation}

\begin{figure}[h]
	\begin{center}
	\includegraphics[scale=0.6]{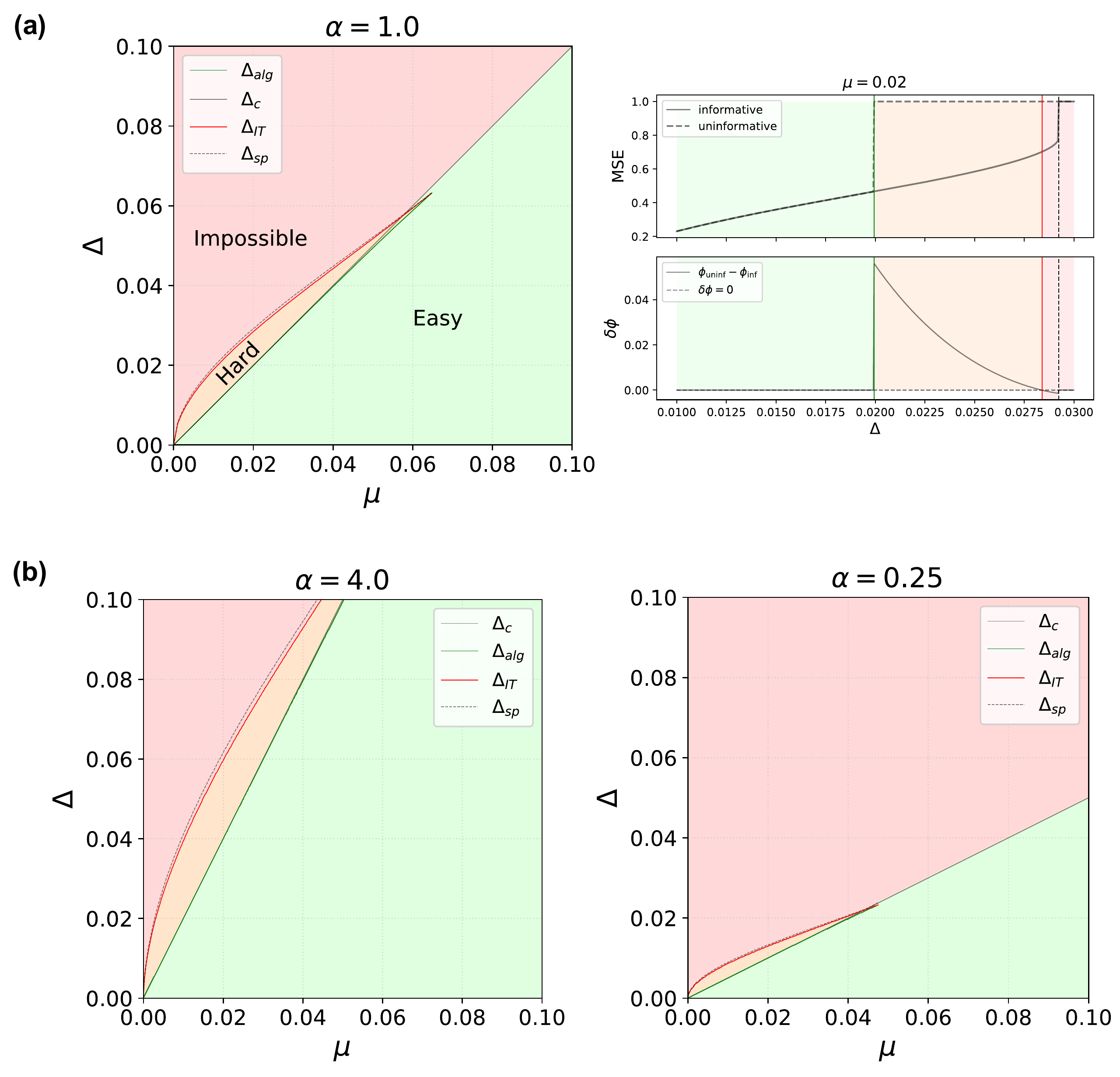}
	\end{center}
	\caption{(a) left panel: The phase diagram for a Rademacher-Bernoulli
          prior on $\theta$ with $\lambda=1/2$ and no bias in the
          distribution of the labels ($\beta = 1/2$). When the
          fraction of spammers is very large (small $\mu$) a hard
          phase appears where the AMP algorithm is not able to reach
          the information-theoretically optimal performance. (a) right
          panel: cut of the phase diagram corresponding to
          $\mu=0.02$, i.e. only 2\% of non-spammers. We plot the MSE (top)
          and the difference in the free energy (bottom) of the two fixed
          points as a function of $\Delta$. Note that in this case we
          still have $\Delta_{\text{c}} > \Delta_{\rm alg}$ but both are indistinguishably
          close. In the hard region (orange) the AMP algorithm reaches
          $\text{MSE}=1$ while the Bayes-optimal estimator reaches the
          depicted MSE. 
          (b) Phase diagrams with all parameters set to the same values, but $\alpha$ different. 
          When $\alpha$ grows (shrinks) inference becomes easier (harder) 
          and the hard region grows (shrinks). The tricritical point for $\alpha=1/4$ is 
          located around $\mu\approx 0.048$ whereas for $\alpha=4$ it is around $\mu\approx 0.077$.
          }
	\label{fig:2D_RB}
	\end{figure}

In Fig.~\ref{fig:2D_RB} we present the phase diagram for several
values of $\alpha=M/N$. We plot the stability threshold $\Delta_{\text{c}}$
as well as the three phase transitions associated with the existence
of the hard phase. We mark the phases where inference is
algorithmically easy, hard and impossible. In particular, we find that
a \emph{hard phase appears} for small enough $\mu$ as depicted in the
figure. Regions with small $\mu$ correspond to crowds that
contain mostly spammers. For $\alpha=1$ the hard phase appears only 
if the vast majority of the workers are spammers. 
When $\alpha$ grows (shrinks) the hard region grows (shrinks) as well.
In the region where the hard phase is absent 
(\ref{eq:stability_criterion_RB}) provides the right criterion to
locate the phase transition from the easy to the impossible phase.

\subsection{Biased labels and worker reliabilities}

\begin{figure}[!ht]
	\begin{center}
	\includegraphics[scale=0.5]{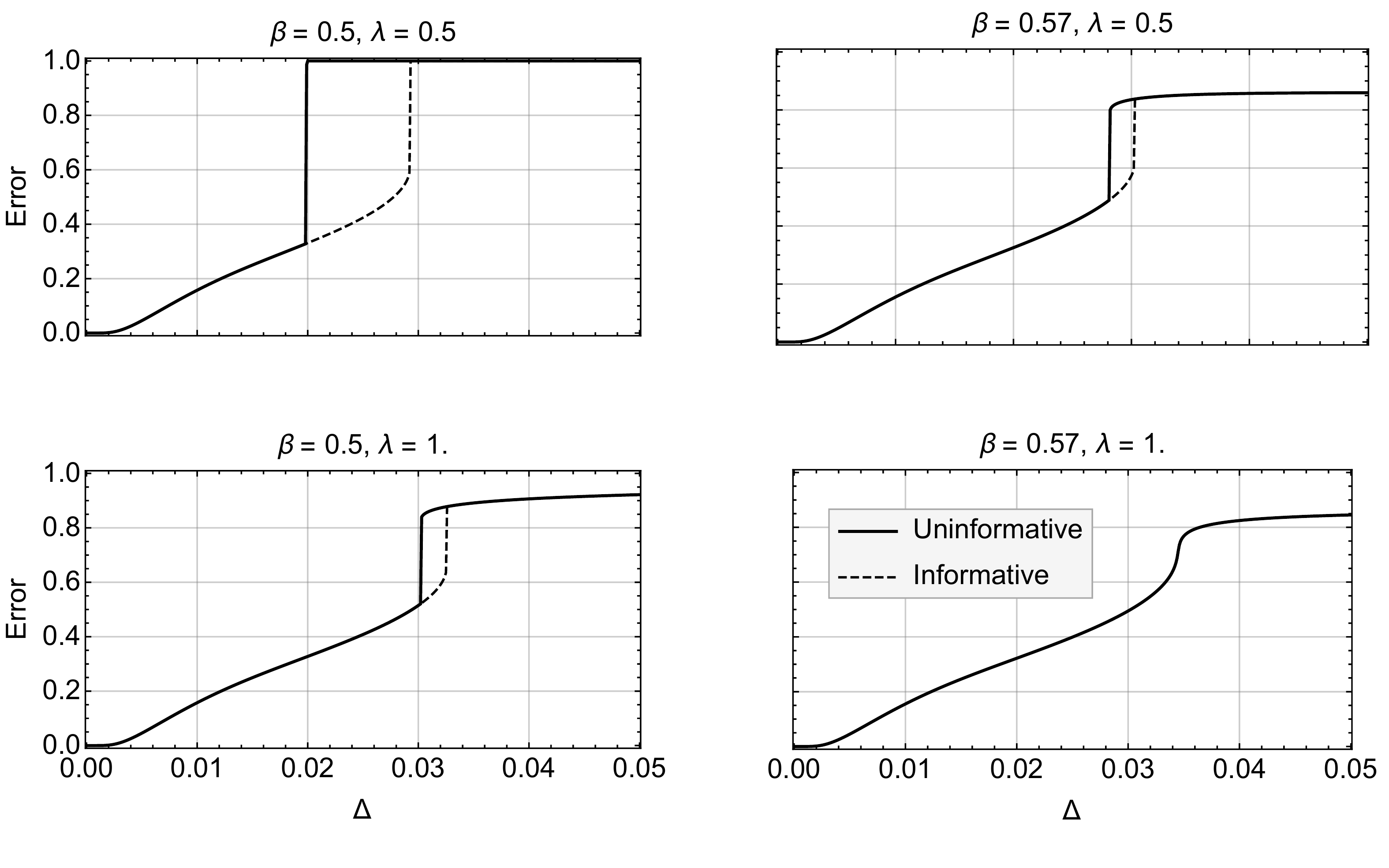}
	\end{center}
	\caption{Influence of bias in the distribution of labels and
          worker reliabilities on the performance. 
          Here we plot the resulting error rate (\ref{eq:ER}) for
          $\alpha=1$ and $\mu=0.02$ as reached from the uninformative
          (bold) and informative (dashed) initialization. 
	For bias in the labels ($\beta \neq 0.5$) or in the workers
        abilities ($\lambda \neq 0.5$) the trivial fixed point (error
        equal to one) is replaced by another fixed point with slightly
        lower error. 
	The hard phase in these examples appears at larger noise and shrinks 
	or might disappear as in the bottom right panel.
	}
	\label{fig:1D_RB_parametric}
	\end{figure}

If $\lambda \neq 1/2$ or $\beta \neq 1/2$ the trivial fixed point $M_v=M_\theta=0$ does
not exist anymore. We illustrate in Fig.~\ref{fig:1D_RB_parametric} how
this changes the phase diagram and the achievable MSE. For the 
case $\alpha=1$ and $\mu=0.02$ we plot the MSE reached by
the state evolution from the informative and the uninformative initialization. 

First (left top panel), we consider the unbiased case with $\beta =
1/2$, but $\lambda \neq 1/2$ as already plotted in Fig.~\ref{fig:2D_RB}. 
In the bottom-left panel we consider the case where $\lambda$ changes. 
Due to the present symmetry it suffices to restrict the attention to $\lambda > 1/2$. 
When more hammers than adversaries are present, i.e.~for $\lambda > 1/2$ the trivial fixed point at $M_v = 0$ 
disappears and instead another fixed point with low but positive overlap (i.e.~error
smaller than 1) appears. The hard phase shrinks as shown in the 
bottom-left panel of Fig.~\ref{fig:1D_RB_parametric}.

If the dataset is biased, i.e. $\beta \neq 1/2$, the
change is quantitatively more dramatic, but phenomenologically very
similar, cf.~top-right panel in Fig.~\ref{fig:1D_RB_parametric}. 
Upon slight change in $\beta$ the hard phase shrinks considerably. 
For a large range of values of $\beta$ and $\lambda$ the hard phase
entirely disappears as in the bottom-right panel in Fig.~\ref{fig:1D_RB_parametric}. 

\subsection{The impact of $\alpha$} 

\begin{figure}[!ht]
	\begin{center}
	\includegraphics[scale=0.45]{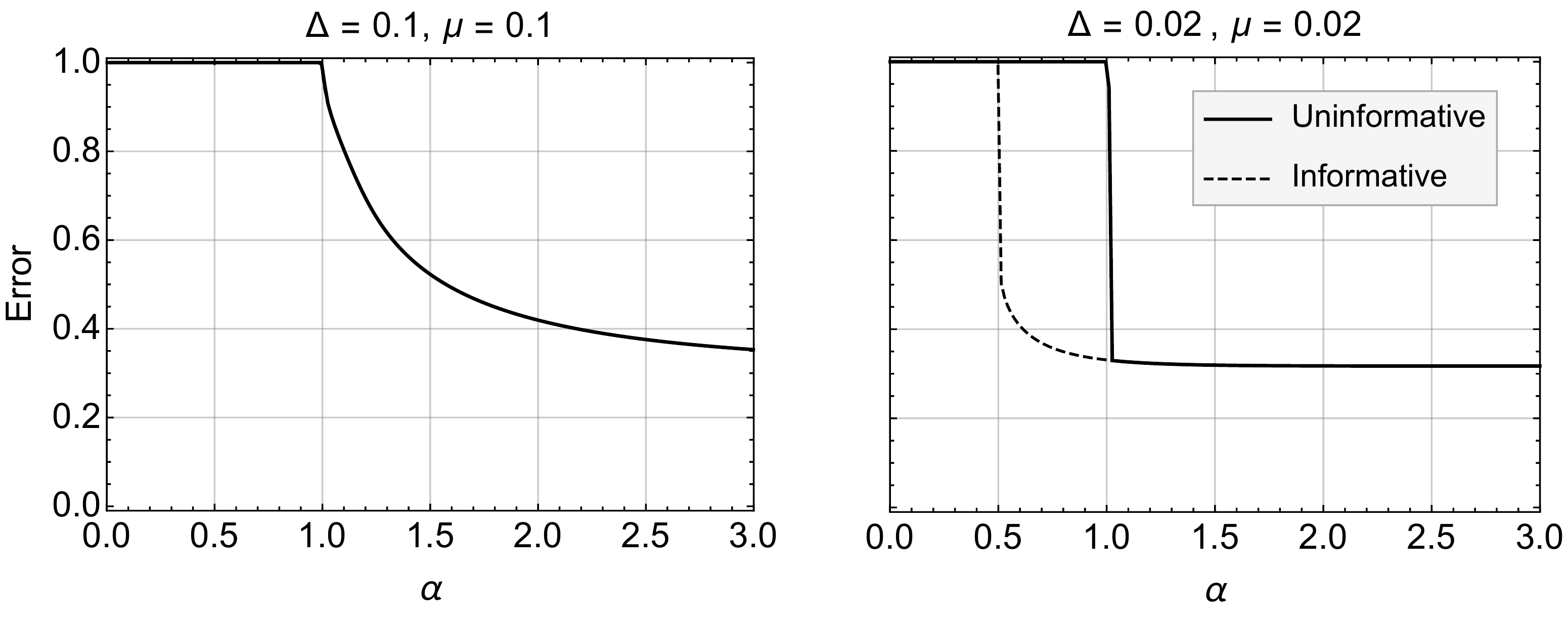}
	\end{center}
	\caption{
	The behaviour of the error rate versus $\alpha$ for the RB prior (\ref{eq:rb_prior}) 
	with $\lambda=1/2$ and bias is set to $\beta=1/2$. 
		}
	\label{fig:alpha}
	\end{figure}

Recall that $\alpha$ is the ratio of tasks to workers in our model. 
By virtue of the $\sqrt{\nu / N}$ scaling of the signal,
cf.~(\ref{eq:channel}), we have two competing mechanisms when $N$ is
increased: on the one hand the signal becomes weaker, on the other
hand we obtain more answers per question. 
Equation (\ref{eq:stability_criterion_RB}) tells us
that we should expect inference to become easier when $\alpha$
increases. Indeed, if we fix $\Delta$ and consider how the performance
changes with $\alpha$ it follows from the SE
that in order to achieve higher overlap 
it is necessary to increase the fraction of questions 
distributed to each worker, i.e.~by increasing $\alpha$. 
This improves the estimation of $\boldsymbol{\theta}$, 
which in turn improves the estimate of $\mathbf{v}$.
We depict this by plotting the error rate against $\alpha$ 
for two different values in Fig.~\ref{fig:alpha}.

How does the hard phase vary with $\alpha$? We answer this question in Fig.~\ref{fig:2D_RB} (b) where we show that the hard phase grows further in the impossible phase when $\alpha$ is increased, while it shrinks when $\alpha$ is decreased.

\subsection{Dealing with other priors}

The derivation of section \ref{sec:model} applies to any prior as long as $\theta=O(1)$. Indeed, many features persist if we replace (\ref{eq:rb_prior}) by 
	\[
	P_\theta (\theta) = (1-\mu) \delta(\theta) + \mu \phi(\theta)
	\, .
	\] 
with $\phi(\theta)$ some appropriate distribution (we have considered $\phi(\theta)$ being a beta distribution or a Gaussian). For instance (\ref{eq:linear_RB}) still holds when $\phi(\theta)$ is a standard Gaussian and as for the RB prior a first order transition is triggered by very noisy $\theta$, i.e. only very few hammers and mostly spammers in the crowd. 

One might also replace the delta distribution by some other sparsity inducing distribution.  
A case for which the corresponding integrals are tractable analytically is that of a mixture of two Gaussians, centered around $\bar\theta_L$ ($\bar\theta_R$) with variance $\sigma_L^2$ ($\sigma_R^2$).
	\[
	P_\theta (\theta) = (1-\mu) \, \mathcal{N}(\theta;\bar\theta_L,\sigma_L^2)+ \mu \, \mathcal{N}(\theta;\bar\theta_R,\sigma_R^2)
	\, .
	\]
Under this choice and with $\beta = 1/2$ in (\ref{eq:prior_v}) the SE equations (\ref{eq:state_evolution}) can be expressed as
	\begin{eqnarray}
	M_v^{t+1} = G \left(\frac{1}{\Delta} T\left( \frac{\alpha}{\Delta} M_v^{t} \right) \right)
	\end{eqnarray}
with 
	\begin{align*}
	G(x) &= \mathbb{E}_W \left\{ \tanh(x + \sqrt{x} W) - \tanh(-x + \sqrt{x} W) \right\}
	\\
	T(q) &= \mu \cdot \mathbb{E}_W \left\{
	\frac{ \left[
	\left( \bar\theta_R +\sqrt{\frac{q}{1+q\sigma_R^2}} \sigma_L^2 W \right)
	+
	\frac{1-\mu}{\mu} \left( \frac{1+q\sigma_R^2}{1+q\sigma_L^2} \right)^{\frac{3}{2}}
	\left( \frac{ \bar\theta_L +q\sigma_L^2 \bar\theta_R }{1+q\sigma_R^2}+\sqrt{\frac{q}{1+q\sigma_R^2}} W \right)
	\cdot \exp\left( -\frac{1}{2} Q(W) \right)
	\right]^2
	}{
	1+\frac{1-\mu}{\mu} \sqrt{\frac{1+q\sigma_R^2}{1+q\sigma_L^2}} \exp\left( -\frac{1}{2} Q(W) \right)
	}
	\right\}
	\\
	Q(W) &= \frac{1+q\sigma_R^2}{1+q\sigma_L^2} \left( W+\sqrt{\frac{q}{1+q\sigma_R^2}} (\bar\theta_R-\bar\theta_L) \right) - W^2
	\, ,
	\end{align*}
where $\mathbb{E}_W$ indicates the average over the standard Gaussian measure on $W$. 
Varying the means ($\bar\theta_L$, $\bar\theta_R$) and variances
($\sigma_L^2$, $\sigma_R^2$) then allows to interpolate between
different scenarios.

\section{Relevance of the results in the sparse regime \label{sec:sparse_limit}}

Our analysis of the dense DS model is based on the ground that the
underlying graphical model (the bipartite question-worker-graph) is
densely connected. This means that each task-node is connected 
to $\Theta(N)$ worker-nodes -- and reversely each 
worker-node is connected to $\Theta(M)$ task-nodes. 
Allowing that some of the tasks remain unanswered 
introduces a sense of sparsity in the channel (cf. (\ref{eq:channel})). 
Our analysis assumes that $1-\rho =\Theta(1)$. 
Existing mathematical literature on low-rank matrix estimation
shows that the formulas we derived for the Bayes-optimal performance,
hold true even when the degrees in the graph grow with $N$ slower than
linearly, i.e.~when $(1-\rho)N$ diverges with $N \to \infty$
\cite{deshpande2015asymptotic,caltagirone2017recovering}. The regime
where the above asymptotic results do not hold anymore is when $1-\rho
= O(1/N)$, which we refer to as the sparse regime.
In this section we investigate numerically 
how the behaviour of the sparse DS model deviates from the 
predictions drawn from the dense DS model. 

In the sparse regime considered here every worker is connected to
$d$ randomly chosen tasks, where $d=\Theta(1)$. Unless the quality
of each answer is very high, the effective noise
$\Delta=[(1-\rho)\nu]^{-1}$ is overwhelming and inference impossible,
unless $\nu=\Theta(N)$. Therefore we will consider the following ``mapping''
	\begin{equation}
	\rho = 1-\frac{d}{M} \hspace{1cm} \nu = n \cdot N
	\, ,
	\label{eq:sparse_analogy}
	\end{equation}
with $n\in [0,1]$ being a constant. Consequently in the sparse regime
we are dealing with high quality workers as compared to the dense regime. 
This brings us close to the setting of previous literature on the DS model 
\cite{KargerOhShah11,Liu12,ok16,ok2016optimal}.

\subsection{Approximate message passing on sparse graphs \label{subsec:sparse_amp}}

We study numerically how the AMP algorithm behaves
when the average degree of the nodes is small. 
In the following we will set $M=N$ such that the average degree 
of the task-nodes equals the average degree, $d$, of the worker-nodes. 

\begin{figure}[H]
	\begin{center}
	\includegraphics[scale=0.5]{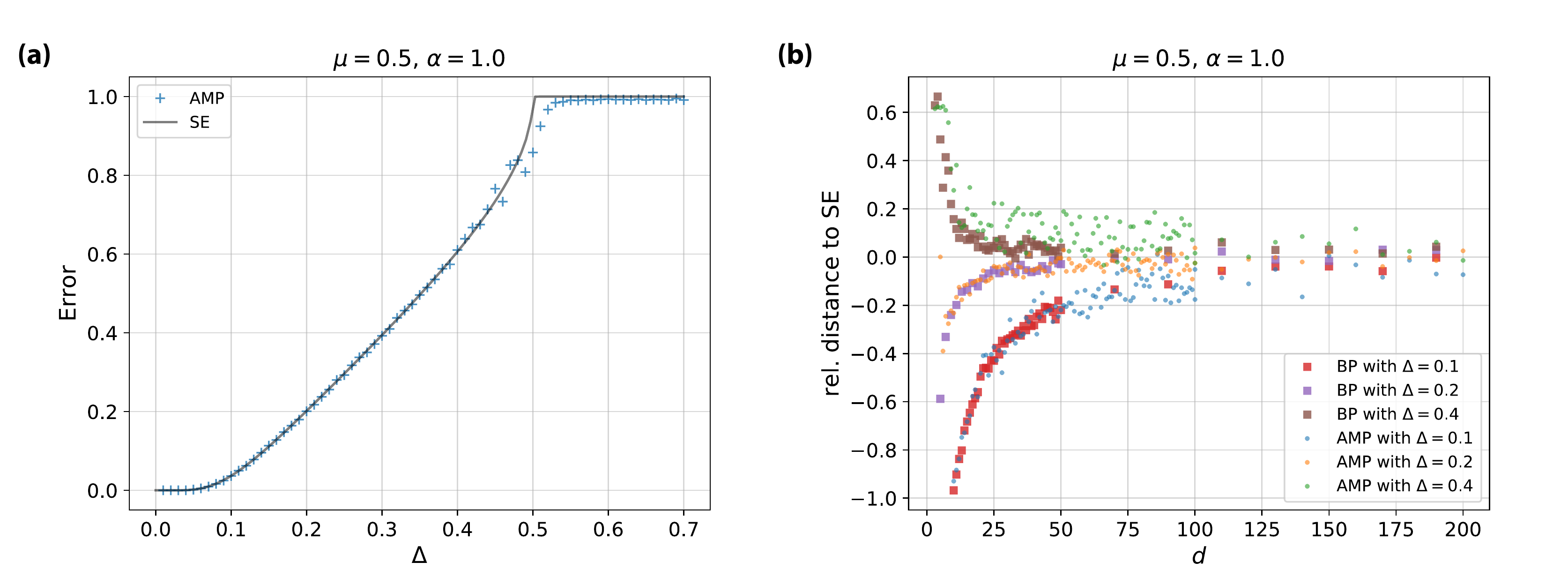}
	\end{center}
	\caption{ 	\label{fig:sparse_amp_1}
	Evaluation of the AMP results in the sparse regime.
	(a) Numerical results for AMP in the dense regime for $N=M=10^4$, averaged over $20$ samples.
	(b) The relative distance of the AMP results to the SE
        prediction of the error when the average degree $d$ and signal
        to noise ratio $\nu$ are varied such that $\Delta$ remains
        fixed. We also compare to the BP algorithm that is asymptotically
        exact in the sparse regime. We see that the SE
        gives an accurate description, already for $d$ around $30-50$. While AMP
        is suboptimal for low degrees $d$ and BP still asymptotically
        optimal, we see that AMP and BP give comparable 
        results down to average degrees around $10$.
        }
	\end{figure}

Figure \ref{fig:sparse_amp_1} (a) depicts results that are obtained by
running AMP in the dense regime where $d=\Theta(N)$, 
for a system with $10^4$ nodes. 
Except from finite size effects close to the phase transition 
the SE prediction agrees with the empirical results.
For Fig.~\ref{fig:sparse_amp_1} (b) we fixed different values of
$\Delta$ -- by adjusting $n$ so that $\Delta=\alpha/(nd)$ -- and
plotted the relative deviation from the SE when the degree $d$ is
varied. We also show the results obtained with the BP algorithm of
\cite{Liu12} that are obtained by matching the prior and signal to
noise ratio. In the limit of large $N$ the BP results are exact even
for finite $d$. We find as expected that when $d$ is increased, the AMP performance
approaches the prediction of the associated dense model and so does
BP. While for very small $d$ BP slightly outperforms AMP, the
difference is not very significant (up to fluctuations).

\begin{figure}[H]
	\begin{center}
	\includegraphics[scale=0.5]{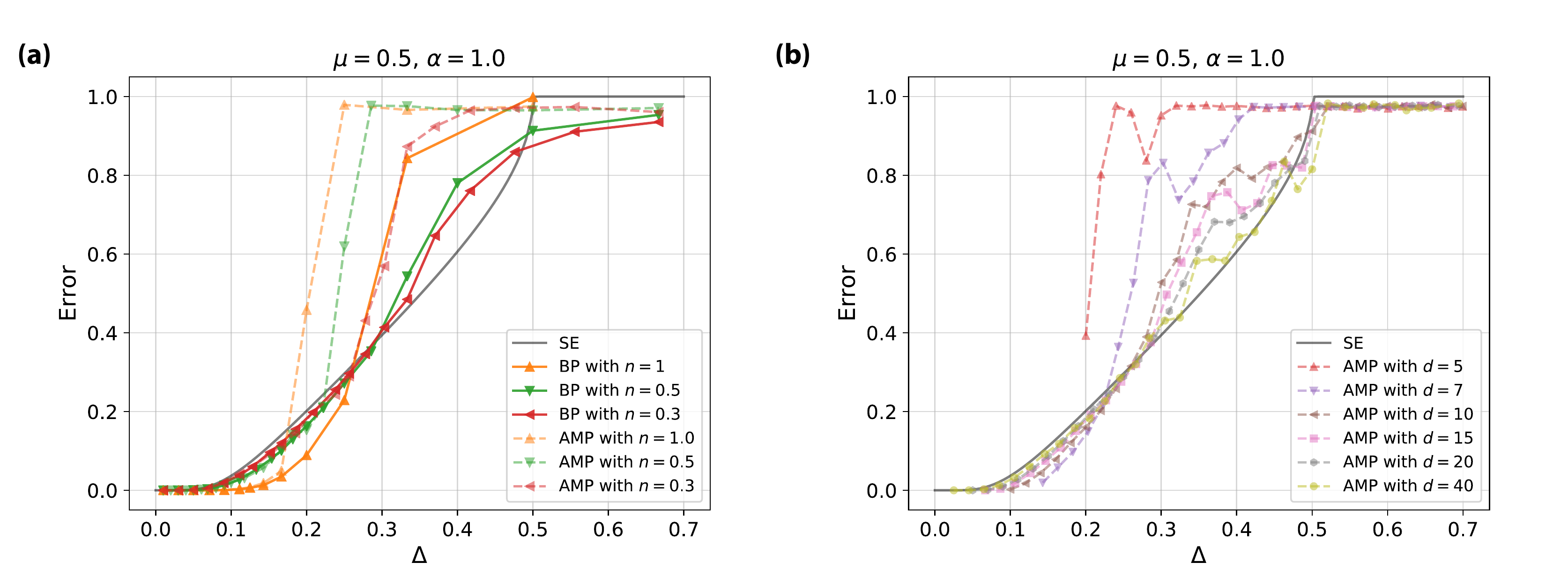}
	\end{center}
	\caption{
	The effect of variation of either $d$ or $\nu$ (i.e.~$n$) on
        the performance of AMP.
	(a) AMP results for fixed $\nu$. We also compared to the BP 
	results that have the same prior and matching signal to noise ratio.
	(b) AMP results for fixed $d$. 
	The fact that the error found in the experiments for large $\Delta$ 
	is slightly below the SE is due to finite size effects. 
	Increasing the average degree pushes the results closer to the SE prediction.
	The experiments were carried out with $N=10^3$ and are averaged over $100$ samples.
	}
	\label{fig:sparse_amp_2}
	\end{figure}

We further quantify the difference in performance of BP and AMP in the
sparse regime in Fig.~\ref{fig:sparse_amp_2}. Now $\nu$ (i.e.~$n$)
is fixed and $d$ (and hence $\Delta$) varies. We compare AMP with its
BP equivalent and find that BP always outperforms AMP, but again only slightly. 
The general trend is as expected: in the sparse regime BP is optimal 
and no other algorithm can outperform it. 
However, it is remarkable how quickly AMP becomes comparable to BP.
In Fig.~\ref{fig:sparse_amp_2} (b) we fix $d$ and vary $\nu$
(i.e.~$n$), such that $\Delta$ varies in the same range as 
in Fig.~\ref{fig:sparse_amp_2} (a). We cannot explore the full range of
$\Delta$ because we must restrict $n\leq 1$.

The results clearly suggests that (for finite size systems) AMP can
indeed be run even on relatively sparse instances. Compared to BP
it is algorithmically less complex and more memory efficient, as
fewer messages need to be stored. 
Further, the state evolution prediction seems to remain a good 
qualitative approximation to the algorithmic performance. 
It suggests that the phenomenology found in the dense limit 
should be rather generic and also appear in sparse systems.

\subsection{First order phase transition in belief propagation \label{subsec:BP}}

So far we have shown that the dense DS model can exhibit both second
and first order phase transitions. The first order transitions are
more interesting algorithmically as they are associated with the
presence of an algorithmically \emph{hard} region where the
corresponding message passing algorithm is suboptimal. 

The authors of \cite{ok16,ok2016optimal} established that 
BP is optimal in the sparse DS model for sufficiently large
signal-to-noise-ratio.
It remains to be tested whether we can observe 
a first order phase transition also in the
sparse version of the model. 
This shall be the aim of the present section.

\begin{figure}[H]
	\begin{center}
	\includegraphics[scale=0.8]{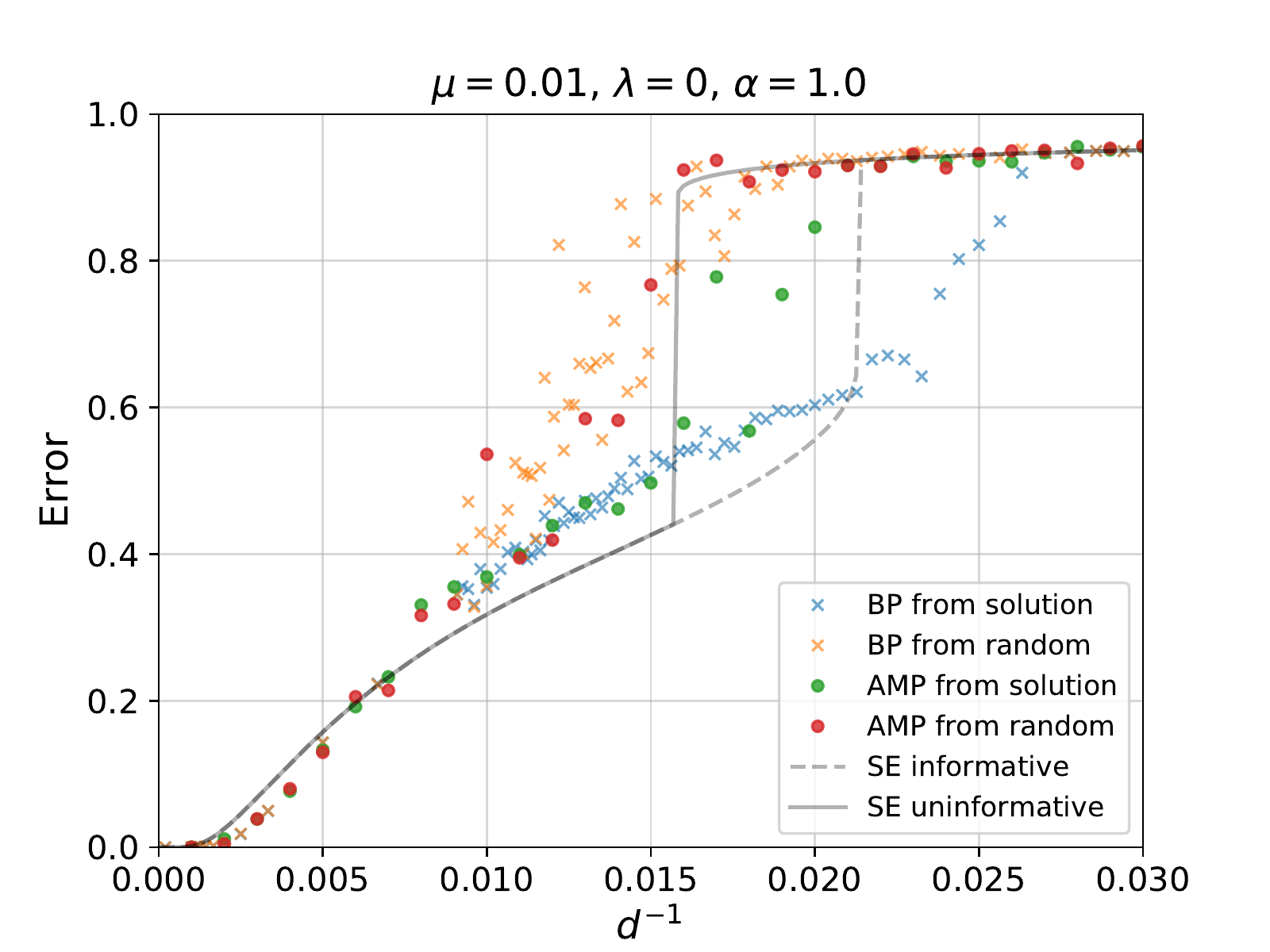}
	\end{center}
	\caption{Numerical results obtained for the BP algorithm of
          references \cite{Liu12,ok16}. The experiments were carried
          out on graphs of size $N=M=10^4$, and are reported as a function
          of the inverse average degree of the worker nodes $d$. A region
          of coexistence associated to a first order phase transition
          opens up and an informative initialization leads to another
          fixed point than the uninformative one. This makes BP sub-optimal
          in the part of this region, where the free energy of the
          fixed point reached from the uninformative
          initialization is higher than the one of the
          fixed point reached from the informative initialization. 
	We found in our experiments that the first order transition appears 
	more pronounced the larger the system size, 
	suggesting that the phenomenon persists asymptotically.}
	\label{fig:sparse_BP}
	\end{figure}

Suboptimality of BP is associated with a region of parameters for
which BP converges to different fixed points from the informative and
from the uninformative initialization. We use our intuition based on the 
results of the dense case to show that there exist regimes where BP is sub-optimal.
Figure \ref{fig:sparse_BP} depicts numerical results obtained for BP with a Bernoulli-prior on $\boldsymbol{\theta}$ ($\lambda=0$ in (\ref{eq:rb_prior})) with very sparse signals ($\mu=0.01$). 
We also plot the AMP performance (in the same sparse regime) as well as the asymptotic prediction 
that would be expected in the dense case (\ref{eq:sparse_analogy}).
Indeed, a clear first order transition appears.
This establishes the suboptimality of BP by virtue of the dependency on the initialization.

\subsection{Approximate message passing on real data \label{subsec:real_world}}

We tested the AMP algorithm (Alg.~\ref{alg:AMP}) on the bluebird dataset of Welinder et al. \cite{Welinder10}. This dataset is fully connected, minimizing effects introduced by poorly designed task-worker-graphs.
We used the same priors and parameters as in \cite{Liu12} to compare AMP to other algorithms. 
Following \cite{Liu12} we also implemented a ``two-coin'' extension of AMP that assumes that the true positive and true negative rates are different. We define $\vec{\theta}_i= (s_i, t_i)$ with $s_i$ the sensitivity of worker $i$ and $t_i$ indicating its specificity. We have
	\begin{align*}
	P\left( Y_{ij} = \pm 1 \mid \vec{\theta}_i ,\  v_j =+1 \right) &= (1-\rho) \cdot \frac{1}{2} \cdot \left( 1 \pm \sqrt{\frac{\nu}{N}} \, s_i \right)
	\\
	P\left( Y_{ij} = \pm 1 \mid \vec{\theta}_i ,\  v_j =-1 \right) &= (1-\rho) \cdot \frac{1}{2} \cdot \left( 1 \mp \sqrt{\frac{\nu}{N}} \, t_i \right)
	\\
	P\left( Y_{ij} = 0 \mid \vec{\theta}_i  ,\  v_j \right) &= \rho
	\, .
	\end{align*}
As in section \ref{subsec:low_rank} we cast the above model into a 
rank-$2$ matrix factorization problem by setting 
	\[
	\vec{v_j} = 
	\begin{pmatrix}
	1\\
	0
	\end{pmatrix}
	\ 
	\text{if question $j$ is true}
	\text{ and }
	\vec{v_j} = 
	\begin{pmatrix}
	0\\
	-1
	\end{pmatrix}
	\ 
	\text{if question $j$ is false}
	\, .
	\]
The only difference is that the former rank-$1$ matrix $\mathbf{w}$,
cf.~(\ref{eq:low_rank_matrix}), now becomes a rank-$2$ matrix with
$\boldsymbol{\theta}\in \mathbb{R}^{M\times 2}$ and $\mathbf{v}\in
\mathbb{R}^{N\times 2}$. The equations for a general rank are
derived and given in detail in \cite{LesieurKrzakalaZdeborova17}. 

	\begin{figure}[H]
	\begin{center}
	\includegraphics[scale=0.8]{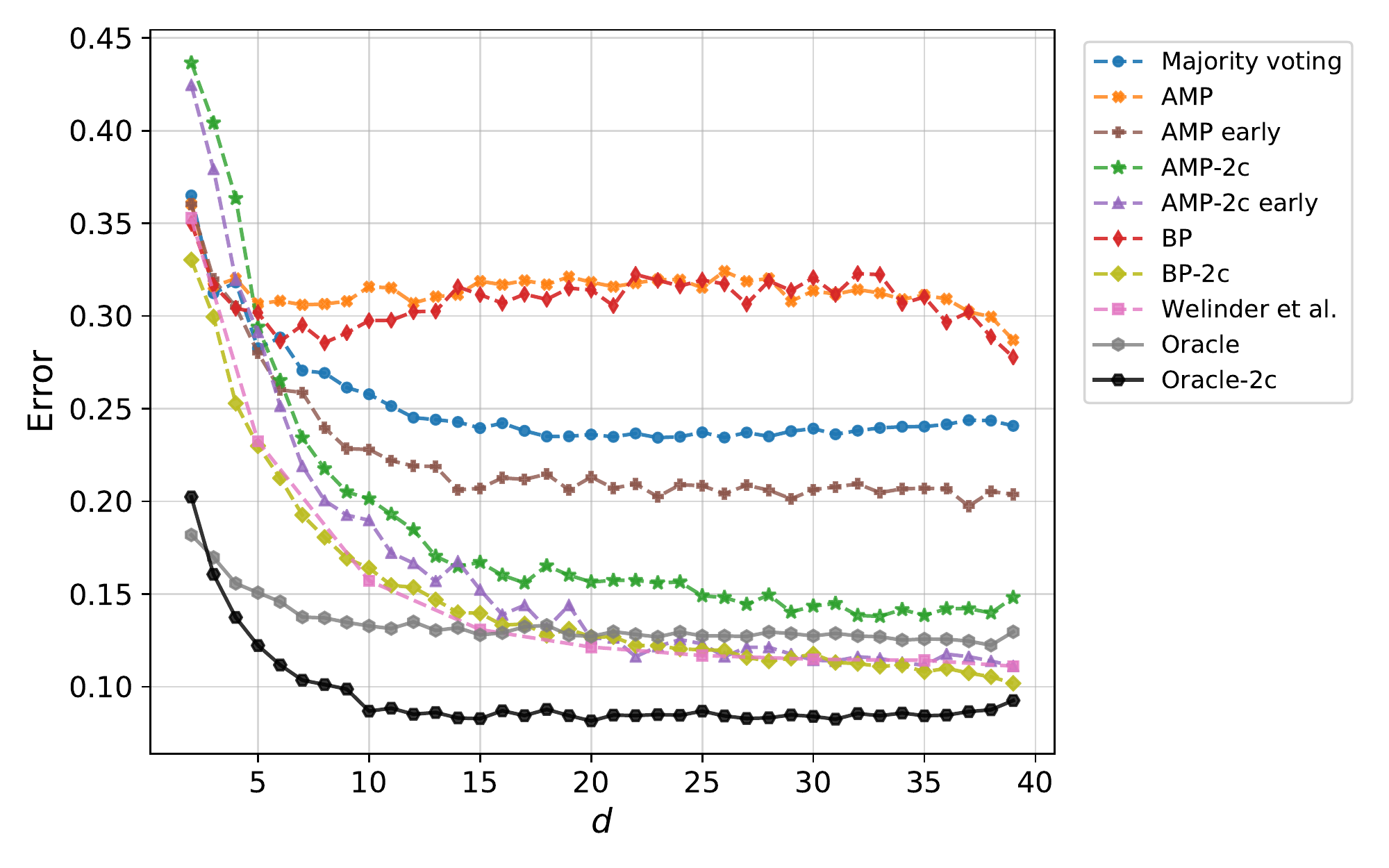}
	\end{center}
	\caption{The error against the number of workers-per-task,
          $d$. Different algorithms are compared to AMP on the
          bluebird dataset. We compare to the results obtained with
          BP, majority voting and the algorithm proposed by Welinder
          et al. \cite{Welinder10}. As explained in the text, we
          implemented two different version of AMP and BP: a
          symmetric one in which the sensitivity and specificity are
          equal and an asymmetric version (referred to as ``2-c'' in
          the legend). Finally we also plot results obtained when AMP
          is run with an early stopping criterion of $10$
          iterations. For BP and AMP the priors are set to independent
          $\text{Beta}(2,1)$ distributions on
          $\boldsymbol{\theta}$. We averaged over $100$ samples for each $d$. 
	}
	\label{fig:real_world}
	\end{figure}

In Fig. \ref{fig:real_world} we compare AMP with BP, majority voting
and the algorithm developed by Welinder et al. in
\cite{Welinder10}. We also compute the oracle lower bound of
\cite{KargerOhShah11} for the two versions of AMP and BP. To evaluate the oracle
it we first estimate the true parameters $\boldsymbol{\theta}$ from
the ground truth and then compute the resulting Bayes-optimal
estimator that maximizes the posterior probability. Note that the
latter estimator has full information of the workers reliabilities. 

We stress that analogous comparisons between existing algorithms and BP
were already performed in \cite{Liu12}, where BP was found to be superior. Our main
point in this section is that AMP, which is simpler than BP, gives a comparable
performance to BP even on real-world data. 
We therefore focus on the comparison between BP and AMP.
Both, BP and AMP perform badly when the original model with $s_i=t_i$ is used 
as can be seen from Fig.~\ref{fig:real_world} by comparing them to majority voting as a baseline algorithm.
Running the same experiments with the two-coin version improves the
results significantly. Indeed BP and AMP perform essentially as well 
as the much more involved algorithm of \cite{Welinder10}.

The experiments were run with identical beta-priors for BP and AMP
($a=2$, $b=1$) for comparability with the results in
\cite{Liu12,ok2016optimal}. For AMP different strategies were
implemented for the prior on $\mathbf{v}$. Setting $\sigma$ to the
true value (estimated from the ground truth) or to $1/2$ led to comparable results as when it was learned. 
In our AMP implementation we initialize $\hat{\mathbf{v}}$ in the estimates obtained by majority voting.

In the symmetric case BP and AMP are very close in performance. The difference for the two-coin models tends to be slightly larger, while the general trend persists. We also observe that it can be beneficial to implement AMP with an early stopping criterion as depicted in Fig.~\ref{fig:real_world}.
Early stopping can be reasonable because the
assumptions made in the derivation are likely to be imprecise,
especially for small system sizes.

In summary, AMP performs quite well on real world datasets. 
The vanilla implementation yields slightly worse results, 
as compared to BP. However, when AMP is stopped after 
few iterations (we used $10$) it reaches much better performance 
in the rank-$1$ case. A significant improvement is also 
obtained in the rank-$2$ version of AMP: for small $d$ BP outperforms AMP, 
but they soon become quasi indistinguishable.
Besides its good performance it has the great advantage 
of algorithmic simplicity, better time complexity and scalability.

\section{Conclusion \label{sec:conclusion}}

In this paper the dense limit of the Dawid-Skene model 
for crowdsourcing was considered.
It was shown that the problem can be mapped onto a 
larger class of low-rank matrix factorization problems. 
This leads to an approximate message passing algorithm 
for crowdsourcing and a closed-form asymptotic analysis of its performance. 
Due to the previous work of \cite{LowRankProof16,MiolaneUV17} 
this analysis can be considered rigorous. While the theory only 
holds rigorously for the dense Dawid-Skene model, numerical
experiments suggest that in the sparse regime AMP still performs well 
and also the asymptotic analysis provides a good qualitative prediction. 

When the crowd consists mainly of spammers with only few workers
that provide useful information, we found that a first order transition appears
in the Bayes-optimal performance. 
Algorithmically this first order transition translates into the
presence of a hard phase in which the AMP algorithm is sub-optimal. 
As a proof of concept we showed numerically that this feature persists
even in the sparse regime where the rigor of our analysis breaks
down. In experiments we also found instances of first order
transitions in the belief propagation algorithm of \cite{Liu12}. This
shows that there are regimes in the Dawid-Skene model where
BP is not optimal. 
This complements recent results on \cite{ok16,ok2016optimal} 
about regimes of optimality of BP. 

We also carried out experiments on real-world data and showed that AMP
performs comparable to other state-of-the-art algorithms, while being
of lower time complexity. Our experiments on the real-world dataset also show
that having a model that described data accurately is more important
than the precise algorithm that is used to do do inference on the model.

\section*{Acknowledgement}

We would like to warmly thank T. Lesieur for advice and guidance as well as Florent Krzakala for suggesting to look at a real world dataset.
LZ acknowledges funding from the European Research Council (ERC) under the European Union'€™s Horizon 2020 research and innovation program (grant agreement No 714608 - SMiLe). This work is supported by the ``IDI 2015'' project funded by the IDEX Paris-Saclay, ANR-11-IDEX-0003-02.

\bibliography{dds.bib}

\end{document}